\documentclass{article}

\usepackage[preprint]{neurips_2026}

\usepackage[utf8]{inputenc} 
\usepackage[T1]{fontenc}    
\usepackage{hyperref}       
\usepackage{url}            
\usepackage{booktabs}       
\usepackage{amsfonts}       
\usepackage{nicefrac}       
\usepackage{microtype}      
\usepackage{xcolor}         
\usepackage{amsmath}
\usepackage{amssymb}
\usepackage[most]{tcolorbox}
\tcbuselibrary{breakable}
\usepackage{float}
\title{Eta Given Delta: Defining LLM Tool Efficiency With Marginal Tool Utility}

\author{%
  Nyx Iskandar \\
  Foam \\
  \texttt{nyx@foam.ai}
  \And
  Perla Gamez \\
  Foam \\
  \texttt{perla@foam.ai} \\
}

\begin{document}

\maketitle

\begin{abstract}
  This paper introduces \textbf{tool efficiency}, a new quantitative metric to evaluate the rate of \textit{useful} tool calls in an LLM agent trajectory. To ensure that tool efficiency is well-defined, we also introduce \textbf{marginal tool utility}, a new quantitative metric defined per tool call indicating whether a tool is useful or whether it can be safely removed from the tool suite without affecting accuracy while increasing tool efficiency; in this paper, we determine the sign of marginal tool utility for each tool call in a trajectory using LLM-as-a-Judge. While much prior work has been done to develop techniques that improve tool use by LLMs and design evaluation methods measuring efficiency indirectly using accuracy as a proxy, our work is centered on measuring efficiency \textit{directly} via the quantitative metric proposed in this paper in post hoc trajectory analyses. It is our intention that this work contributes to the frontier of LLM evaluation research as a springboard for future benchmark designs and agent harness engineering (specifically with regards to creating lean tool suites) that optimize for metrics that complement but are distinct from accuracy.
\end{abstract}

\section{Introduction} \label{section:introduction}

Large Language Models (LLMs) are increasingly used in real-world tasks such as software engineering \citep{chen2021evaluatinglargelanguagemodels, athiwaratkun2023multilingualevaluationcodegeneration, austin2021programsynthesislargelanguage}, mathematical reasoning and autoformalization \citep{ahn2024largelanguagemodelsmathematical, guo2025unspoken, manem2025sandmath}, and scientific research \citep{Zhuang_2025_llmaspr, chaturvedi2026reliablesafesecurellms}. However, should LLMs remain simply as language modeling artifacts \citep{brown2020languagemodelsfewshotlearners} without access to external up-to-date knowledge \citep{cheng2024dated} and no way to act or modify its environment \citep{yao2020calmexplorelanguagemodels, huang2022languagemodelszeroshotplanners, ahn2022icanisay}, they can hardly be called agents \citep{RN2020}. To address this, LLMs are equipped with tools \citep{schick2023toolformerlanguagemodelsteach}, which have become an integral part of the LLM agentic workflows that we see today \citep{yao2022react}.

While tools are generally useful, not all tools are equally useful \citep{huang2024metatoolbenchmarklargelanguage, yu2025chemtoolagentimpacttoolslanguage}, this finding acknowledged even by frontier industry labs through launching features such as tool search \citep{Anthropic2025toolsearch}. To that end, much prior work focuses on developing techniques that improve tool use \citep{sivakumaran-etal-2026-dart} and on creating benchmarks that such techniques can evaluate against and optimize for \citep{ning2024wtuevalwhetherornottoolusage}. A common ground shared by existing work is that the evaluation metric used is almost always outcome accuracy \citep{jin2025searchr, li2025torlscalingtoolintegratedrl}, even if accuracy is not the primary metric of interest, rather it being used as a proxy to measure the efficiency of tool use \citep{ning2024wtuevalwhetherornottoolusage}.

Given this, we notice a gap in the literature: there does not yet exist a direct way to measure tool efficiency, which we define as the rate of useful tool calls, given an LLM agent trajectory. While recent works like TRM by \cite{ma2026empowering} take the first steps towards defining reward functions centered on intermediate tool invocations for reinforcement learning to improve tool call quality, we take this idea one step further by enabling post hoc analyses on agent trajectories with an aggregate tool efficiency metric, and keeping to solely inference-time methods easily adoptable by application-layer engineering teams to assist them in designing leaner and more maintainable agent tool suites.

In attempting to define tool efficiency, we must first define what usefulness means with respect to a tool call. We arrive at the idea of marginal tool utility, which is a metric relevant to each particular tool call instance in an agent trajectory that directly determines the usefulness of these tool calls, and whose aggregate per tool determines the usefulness of the tool itself. This definition must be supported by empirical results, specifically those obtained from tool ablations and observing the change in accuracy (or lack thereof) depending on which tool was ablated. To that end, we used the APEX-SWE Observability benchmark, a benchmark in which today's frontier models still struggle to achieve above $40\%$ accuracy as reported by \cite{kottamasu2026apexswe}, and a benchmark that by default includes tools that we hypothesize (and later prove) are necessary (useful) as well as unnecessary (removable without affecting accuracy).

As such, \textbf{our contributions} in this paper are four-fold:
\begin{enumerate}
    \item Introduce marginal tool utility $\Delta_\alpha$ defined per tool call, specifically for pass/fail tasks. Aggregate tool utility $\sum \Delta_\alpha$ defined per tool predicts whether said tool affects the agent's accuracy when ablated.
    \item Introduce tool efficiency $\eta_\alpha$, where a useful tool call $\alpha_i$ is one with $\Delta_\alpha(\alpha_i) > 0$.
    \item Implement a simple and cost-efficient method for calculating $\eta_\alpha$ for pass/fail tasks by using LLM-as-a-Judge to determine the sign of each $\Delta_\alpha(\alpha_i)$.
    \item Empirically support the correctness of our definitions and LLM-as-a-Judge evaluator through tool ablations on the APEX-SWE Observability benchmark; critically, adding tools with $\sum\Delta_\alpha \leq 0$ does not increase the accuracy of the agent, while adding tools with $\sum \Delta_\alpha > 0$ does increase the accuracy of the agent, all as expected.
\end{enumerate}

\section{Related Works}

\textbf{LLM evaluations.} There exists a plethora of benchmarks to assess an LLM agent's accuracy in completing various tasks, including SWE-bench \citep{jimenez2024swebenchlanguagemodelsresolve}, Terminal-Bench \citep{merrill2026terminalbenchbenchmarkingagentshard}, BrowseComp \citep{wei2025browsecompsimplechallengingbenchmark}, and OSWorld-Verified \citep{OSWorld}. Benchmarks specifically designed to evaluate an agent's ability to use tools to correctly complete tasks include MCP-Atlas \citep{bandi2026mcpatlaslargescalebenchmarktooluse}, ToolBench \citep{qin2023toolllm}, StableToolBench \citep{guo2024stabletoolbench}, and Toolathlon \citep{li2025toolathlon}. While they differ in their exact evaluation method, the aforementioned benchmarks all focus on evaluating accuracy. Even benchmarks to specifically assess the quality of tool use like WTU-EVAL by \cite{ning2024wtuevalwhetherornottoolusage} use accuracy as a proxy to determine whether tool usage is done correctly. The quantitative metrics that we introduce in this paper allow future benchmarks to bypass accuracy altogether, or use them as complementary metrics to accuracy.

\textbf{LLM tool call optimizations.} Many methods have been proposed by existing work to optimize LLM tool usage: reward models \citep{agarwal2026toolrmoutcomerewardmodels, ma2026empowering}, self-play reinforcement learning \citep{acikgoz2026toolr0selfevolvingllmagents}, and tool retrieval \citep{Moura2025toolrag, erdogan2024tinyagentfunctioncallingedge}. Other interesting related work include ART \citep{paranjape2023artautomaticmultistepreasoning} that approach tool call optimization not necessarily from the lens of improving tool usage, but using tool outputs to improve generation dynamically. Several works like \cite{yu2025chemtoolagentimpacttoolslanguage} conclude that some tasks benefit from tool usage while others do not, suggesting that tool usefulness is task-dependent and not equal across all tools. Similar to \cite{ma2026empowering}, we extend this idea and hypothesize that usefulness is not equal across all tool \textit{calls}, though instead of developing reward models for reinforcement learning, we derive aggregate metrics to be used by application-layer developers conducting post hoc analyses on their agent trajectories to optimize their agent harness \citep{lou2026autoharnessimprovingllmagents} without fine-tuning \citep{zhang2024offlinetraininglanguagemodel}. Moreover, marginal tool utility and tool efficiency can add another dimension to future dynamic tool synthesis methods like Test-Time Tool Evolution \citep{lu2026statictoolstesttimetool}, which currently calculates semantic similarity between planning steps and tool descriptions to dynamically select tools from a tool library or to generate and refine new tools.

\textbf{Methods improving agentic efficiency.} To the best of our knowledge, tool efficiency is not a well-defined nor quantitative metric prior to this paper. However, this does not mean that there has been no prior work that is focused on improving efficiency in LLM agentic workflows in the broad sense. For instance, Recursive Language Models \citep{zhang2026recursivelanguagemodels} and FrugalGPT \citep{chen2024frugalgpt} focus on minimizing API cost, LLMLingua \citep{jiang-etal-2023-llmlingua} on minimizing inference latency, and an instruction-refinement framework by \cite{wu-etal-2025-joint} on maximizing Cost-Aware Pass Rate (which is a metric introduced in the same paper). We consider the metrics these methods aim to optimize \textit{efficiency-adjacent}, reserving the term \textit{efficiency} for a metric that calculates the rate of \textit{useful} items as is accepted in physics, where efficiency is the ratio of useful energy to total energy input of a system \citep{Serway_Jewett_Peroomian_2012}.

\section{Methodology} \label{section:methodology}

To enable direct and explicit measurements of efficiency, we first introduce \textbf{marginal tool utility}, particularly in the case of pass/fail tasks. We then introduce \textbf{tool efficiency}, where the usefulness of a tool call corresponds to the sign of marginal tool utility of said tool call.

\subsection{Problem Setup}

We formalize a multi-turn LLM agent trajectory as an ordered sequence of token sequences. Assuming the ReAct paradigm \citep{yao2022react}, which is the paradigm used for the agent harness in APEX-SWE Observability \citep{kottamasu2026apexswe}, an LLM $\pi$ alternates between generating reasoning tokens and executing tool calls.

Formally, consider LLM $\pi$ as well as system message $s$ and user message $u$. Given prompt $p \triangleq (s, u)$, LLM $\pi$ executes multiple tool calls sequentially and outputs reasoning tokens in between tool calls to synthesize the information it has access to at each timestep. This iterative process terminates once LLM $\pi$ reaches the maximum allowable step count or calls a terminal tool to produce final answer $y$. Hence, we define agent trajectory $\tau$ as

\begin{equation}
    \tau \triangleq (p, t_1, r_1, o_1, ..., t_N, r_N, o_N, t_{N+1}, y)
\end{equation}

where each $t_i \space (1 \leq i \leq N+1)$ denotes a sequence of reasoning tokens, each $r_i \space (1 \leq i \leq N)$ denotes a tool request, each $o_i \space (1 \leq i \leq N)$ denotes the tool output corresponding to $r_i$, and $N \in \mathbb{N}$ denotes the total number of steps taken by the agent that is no greater than the maximum step count set as a hyperparameter. We define a tool call as a paired tool request and tool output, or formally,

\begin{equation}
    \alpha_i \triangleq (r_i, o_i)
\end{equation}

To validate our paper's proposed definitions, we must define the expected correctness of final answer $y$. In APEX-SWE Observability, final answer $y$ is the final code patch generated by the agent to solve its automatic program repair task \citep{desouza2017spectrumbasedsoftwarefaultlocalization}. Correctness is determined by unit tests following the $\text{FAIL\_TO\_PASS} / \text{PASS\_TO\_PASS}$ methodology inspired by SWE-bench \citep{jimenez2024swebenchlanguagemodelsresolve}. We define each unit test as function $f_j \space \colon \space \mathbb{N}^n \rightarrow \{0,1\}$, where $f_j(y) = 0$ when final answer $y$ causes unit test $j$ to fail and $f_j(y) = 1$ when final answer $y$ causes unit test $j$ to pass. The expected correctness of final answer $y$ over trajectories generated by LLM $\pi$ is thus

\begin{equation} \label{eq:e-correctness}
    \mathbb{E}_{\tau \sim \pi}[\mathbb{I}(f_1(y) \cdot ... \cdot f_M(y) = 1)]
\end{equation}

where $M \in \mathbb{N}$ denotes the total number unit tests such that $1 \leq j \leq M$, and $\mathbb{I}(\cdot)$ is the indicator function.

\subsection{Defining Marginal Tool Utility} \label{subsection:mtu}

Given trajectory $\tau$ of an agent tasked to solve a pass/fail task, the marginal tool utility of the $i^{\text{th}}$ tool call $\Delta_\alpha(\alpha_i)$ is defined as the difference between the likelihood that the task is solved correctly given tool calls up to and \textit{including} $\alpha_i$ and the likelihood that the task is solved correctly given tool calls up to and \textit{excluding} $\alpha_i$. Among other methods, this can be determined using repeated policy $\pi$ rollouts or using LLM-as-a-Judge \citep{zheng2023judging, yu2025aisjudgeaisrise} comparing the before and after trajectories. The latter is more computationally efficient and practical than the former, hence we opt to use LLM-as-a-Judge in this paper, specifically to determine $sgn(\Delta_\alpha(\alpha_i))$, where $sgn(\cdot)$ is the signum function. We argue that this is an acceptable method as a tool call $\alpha_i$ is \textit{useful} if and only if upon its execution the likelihood of correctness increases $(\Delta_\alpha(\alpha_i) > 0)$. Hence, the additional information afforded by repeated policy $\pi$ rollouts, that being the exact likelihood difference before and after the tool call, is unnecessary for our purposes.

Formally, consider judge LLM $\pi^{(J)}$ as well as system message $s^{(J)}$ and user message $u^{(J)}(\alpha_i, \tau)$. Given prompt $p^{(J)} \triangleq (s^{(J)}, u^{(J)}(\alpha_i, \tau))$, judge $\pi^{(J)}$ generates final answer $y^{(J)}$, which is a structured output that can be represented as a triplet $(L, C, R)$, where $L = \{0,1\}$ denoting two possible marginal tool utility labels (\texttt{positive} or \texttt{non\_positive}), $C = [0,1]$ denoting a float confidence score, and $R = \mathbb{N}^n$ denoting tokens that read as the judge's rationale for its classification.

The judge trajectory $\tau^{(J)}$ is thus simply

\begin{equation}
    \tau^{(J)} \triangleq (p^{(J)}, y^{(J)})
\end{equation}

Given judge $\pi^{(J)}$, the $i^{\text{th}}$ tool call $\alpha_i$, the trajectory up to and excluding the $i^{\text{th}}$ tool call $\tau_{1, ..., i-1}$ (i.e., before trajectory), and the trajectory up to and including the $i^{\text{th}}$ tool call $\tau_{1, ..., i}$ (i.e., after trajectory), we get $y^{(J)}_i$. In particular,

\begin{equation}
    \begin{cases}
        L_i = 1 \iff \Delta_\alpha(\alpha_i) > 0 & \text{tool call is useful}\\
        L_i = 0 \iff \Delta_\alpha(\alpha_i) \leq 0 & \text{otherwise}
    \end{cases}
\end{equation}

Notice that marginal tool utility is defined per tool \textit{call}. For a particular trajectory $\tau$, a \textit{tool} is useful if and only if it has more useful $(\Delta_\alpha(\alpha_i) > 0)$ calls than it has non-useful $(\Delta_\alpha(\alpha_i) \leq 0)$ calls. Equivalently, for all tool calls $\alpha_i$ of a particular tool,

\begin{equation}
    \sum \Delta_\alpha(\alpha_i)
        \begin{cases}
            > 0 & \text{tool is useful}\\
            \leq 0 & \text{otherwise}
        \end{cases}
\end{equation}

We can also define this aggregate for a set of independent trajectories. This is only advisable in practical settings if the independent trajectories share the same tool suite and each corresponds to a task comparable to each other (e.g., each corresponds to an observability task using the same agent harness within the same benchmark).

For reproducibility, we outline the specific system message $s^{(J)}$ and user message $u^{(J)}$ in Appendix \ref{app:judge-mtu}, and note that the particular model used for judge LLM $\pi^{(J)}$ is GPT‑5.4 on Microsoft Azure \citep{OpenAI2026gpt54}.

\subsection{Defining Tool Efficiency} \label{subsection:tool-efficiency}

Having determined $sgn(\Delta_\alpha(\alpha_i)) \space \forall i$, we can define tool efficiency as the ratio of the number of useful tool calls to the total number of tool calls in a trajectory. Formally, for trajectory $\tau$ with $N$ tool calls,

\begin{equation}
    \eta_\alpha(\tau) \triangleq \frac{|\{\alpha_i | \Delta_\alpha(\alpha_i) > 0 \space \}|}{N} \text{, $1 \leq i \leq N$}
\end{equation}

We can thus also define the mean tool efficiency of a set of $L$ independent trajectories

\begin{equation}
    \overline{\eta_\alpha}(\tau^{(1)},...,\tau^{(L)}) = \frac{\eta_\alpha(\tau^{(1)}) + ... + \eta_\alpha(\tau^{(L)})}{L}
\end{equation}

We hypothesize, and empirically show in Section \ref{section:results}, that tool efficiency increases when a non-useful tool is removed from the tool suite. This is not trivially implied by the definition of tool efficiency, most significantly because the total number of tool calls $N$ can vary given a change in the tool suite.

\subsection{Experimental Setup}

We conduct our empirical experiments to show that the concept of marginal tool utility aligns with what we expect to see in practical settings when performing tool ablations. Particularly, if a tool has positive aggregate tool utility, the expected correctness of the agent's final answer increases with the inclusion of the tool in the agent's tool suite; if a tool has non-positive aggregate tool utility, the expected correctness of the agent's final answer remains constant or decreases with the inclusion of the tool in the agent's tool suite. We show that these hypotheses are supported by the empirical correctness results.

We evaluate frontier LLMs against the 25 public observability tasks of APEX-SWE \citep{kottamasu2026apexswe} available on HuggingFace. \cite{kottamasu2026apexswe} noted that this public subset is representative of the whole set of 100 observability tasks, of which 75 are private.

In APEX-SWE Observability, the default agent harness includes native agent tools (bash, search\_files, read\_file, apply\_patch, update\_plan) and three read-only MCP tools exposed via bash: Grafana/Loki for logs, Mattermost for discussions between human developers, and Plane for software issues/tickets/specifications. Being familiar with how observability tasks for production software are often solved, we hypothesize that Grafana/Loki is a useful MCP tool with positive aggregate tool utility, while Mattermost and Plane are non-useful MCP tools with non-positive aggregate tool utility.

To prove that hypothesis, we conducted tool ablations to obtain three variants of the agent harness:

\begin{itemize}
    \item \texttt{default}: Full suite of MCP tools (Grafana/Loki, Mattermost, Plane).
    \item \texttt{grafana}: Only Grafana/Loki MCP is available.
    \item \texttt{no-mcp}: Absolutely no MCP tools are available.
\end{itemize}

Through these ablations, we are able to verify (1) whether the inclusion/omission of Mattermost and Plane affects accuracy, and (2) whether the inclusion/omission of Grafana/Loki affects accuracy. In Section \ref{section:results}, we present the task accuracy results we obtained and cross-reference them with the aggregate tool utility results of the three MCP tools. We did not conduct ablations on the native agent tools as they are either necessary for the task to be completed (e.g., apply\_patch) or for the MCP tools to be executed (e.g., bash).

\textbf{Agent specifications.} We used GPT‑5.3‑Codex on Microsoft Azure \citep{OpenAI2026gpt53codex} and Gemini 3.1 Pro on Google Agent Platform (formerly Vertex AI) \citep{Google2026gemini31pro} to create two independent agent instances. Therefore, we obtained a total of $25 \cdot 3 \cdot 2 = 150$ distinct agent trajectories.

\textbf{Compute specifications.} The proprietary LLMs were accessed via their respective provider APIs and thus run on their respective providers' compute clusters. The experiments were otherwise run on AWS Batch using EC2 CPU workers, each as an isolated containerized job with Docker-in-Docker enabled. Jobs used 16 vCPUs, 110 GB RAM, and a 600 GB gp3 root volume. The AWS Batch compute environment used the \texttt{r7i.4xlarge}, \texttt{r7i.8xlarge}, and \texttt{m7i.8xlarge} instance types.

\section{Results} \label{section:results}

In this section, we report the results we obtained from all $150$ agent trajectories, namely task accuracy, marginal and aggregate tool utility, and mean tool efficiency. We show that task accuracy increases with the inclusion of a tool with positive aggregate tool utility, and tool efficiency increases with the removal of tools with non-positive aggregate tool utility.

\subsection{Task Accuracy} \label{subsection:task-acc}

We find that task accuracy does not significantly change when both Mattermost and Plane are removed (\texttt{default} vs \texttt{grafana}) and that task accuracy noticeably decreases when Grafana/Loki is removed (\texttt{grafana} vs \texttt{no-mcp}), as seen in Tables \ref{table:acc-codex} and \ref{table:acc-gemini}.

\begin{table}[H]
  \caption{Accuracy across 25 trajectories per variant for GPT-5.3-Codex. Each set of 25 tasks were run twice resulting in consistent accuracies.}
  \label{table:acc-codex}
  \centering
  \begin{tabular}{cccccc}
    \toprule
    Variant     & Pass     & Fail & Accuracy \\
    \midrule
    \texttt{default} & 8     & 17 & 0.32     \\
    \texttt{grafana}     & 9     & 16 & 0.36      \\
    \texttt{no-mcp}     & 6     & 19 & 0.24  \\
    \bottomrule
  \end{tabular}
\end{table}

\begin{table}[H]
  \caption{Accuracy across 25 trajectories per variant for Gemini 3.1 Pro. Each set of 25 tasks were run twice resulting in consistent accuracies.}
  \label{table:acc-gemini}
  \centering
  \begin{tabular}{ccccccc}
    \toprule
    Variant     & Pass     & Fail & Accuracy \\
    \midrule
    \texttt{default} & 9     & 16 & 0.36     \\
    \texttt{grafana}     & 9     & 16 & 0.36      \\
    \texttt{no-mcp}     & 5     & 20 & 0.20  \\
    \bottomrule
  \end{tabular}
\end{table}

A breakdown of which specific tasks passed/failed can be found in Appendix \ref{app:acc-breakdown}. Comparing \texttt{default} and \texttt{grafana}, when a task is completed correctly for one it is also generally completed correctly for the other. A similar relationship exists for tasks that were completed incorrectly.

From these results, note that we can qualitatively claim that Grafana/Loki is a useful tool, while Mattermost and Plane are not. This claim aligns with the nature of the tasks: an agent completing an observability task benefits more from reading logs (primary source of information) than from reading discussions or feature specifications (secondary sources of information). Furthermore, the realities of software development mean that these secondary sources may instead mislead the agent towards investigating parts of the codebase that are actually not broken as misinterpretations of the bug by human developers may be reflected in the discussions and feature specifications may be outdated even during feature implementation.

\subsection{Marginal Tool Utility} \label{subsection:mtu}

Using the judge LLM, we first determine the sign of marginal tool utility of each tool call in each trajectory. We also analyze the output rationales from the judge LLM for its classifications. Then, we calculate the aggregate tool utility of each tool across all trajectories grouped by model and variant.

In Section \ref{subsection:task-acc}, we see that Grafana/Loki affects accuracy when added/removed from the tool suite, and that neither Mattermost nor Plane do. We thus expect the following aggregate tool utilities:

\begin{equation}
    \begin{cases}
        \sum_{\text{grafana}} \Delta_\alpha(\alpha_i) > 0 \\
        \sum_{\text{mattermost}} \Delta_\alpha(\alpha_i) \leq 0 \\
        \sum_{\text{plane}} \Delta_\alpha(\alpha_i) \leq 0
    \end{cases}
\end{equation}

Investigating the $50$ trajectories belonging to the \texttt{default} variant ($25$ generated by GPT-5.3-Codex and another $25$ by Gemini 3.1 Pro), the marginal tool utility results in Tables \ref{table:mtu-codex} and \ref{table:mtu-gemini} are consistent with that expectation, supporting our proposed definition and implementation. For \texttt{default} with GPT-5.3-Codex, the aggregate tool utility of Grafana/Loki is $+25$, Mattermost is $-35$, and Plane is $-30$. For \texttt{default} with Gemini 3.1 Pro, the aggregate tool utility of Grafana/Loki is $+5$, Mattermost is $-17$, and Plane is $-28$. For \texttt{grafana}, the aggregate tool utility of Grafana/Loki is $58-29=+29$ with GPT-5.3-Codex and $23-20=+3$ with Gemini 3.1 Pro (no corresponding tables).

In determining the signs of marginal tool utilities, the judge was prompted to output its rationale. The most common rationales to justify $\Delta_\alpha > 0$ classifications are as follows: Grafana/Loki logs exposed concrete failures (exact error logs that hint at the likely fix); more targeted queries for Grafana/Loki logs at later steps narrowed the investigation and provided more focused signals; and Mattermost and Plane context helped orient the agent towards useful conversation and/or requested feature context. To justify $\Delta_\alpha \leq 0$ classifications: generic, irrelevant, or noisy context from logs, discussions, and specifications distracted the agent; malformed queries by the agent wasted steps on retries; and valid queries but those resulting in empty outputs wasted steps on pivoting to valid queries that actually gave non-empty outputs.

\begin{table}[H]
  \caption{Marginal tool utility signs of each tool in the \texttt{default} harness for GPT-5.3-Codex.}
  \label{table:mtu-codex}
  \centering
  \begin{tabular}{ccccc}
    \toprule
    Tool     & $\Delta_\alpha > 0$     & $\Delta_\alpha \leq 0$ & Mean Confidence ($\Delta_\alpha > 0$) & Mean Confidence ($\Delta_\alpha \leq 0$) \\
    \midrule
    Grafana/Loki & 52 & 27 & 0.819 & 0.890 \\
    Mattermost & 7 & 42 & 0.734 & 0.927  \\
    Plane  & 23 & 53 & 0.775 & 0.928  \\
    \bottomrule
  \end{tabular}
\end{table}

\begin{table}[H]
  \caption{Marginal tool utility signs of each tool in the \texttt{default} harness for Gemini 3.1 Pro.}
  \label{table:mtu-gemini}
  \centering
  \begin{tabular}{ccccc}
    \toprule
    Tool     & $\Delta_\alpha > 0$     & $\Delta_\alpha \leq 0$ & Mean Confidence ($\Delta_\alpha > 0$) & Mean Confidence ($\Delta_\alpha \leq 0$) \\
    \midrule
    Grafana/Loki & 26 & 21 & 0.825 & 0.870     \\
    Mattermost & 3 & 20 & 0.673 & 0.930      \\
    Plane  & 3 & 31 & 0.607 & 0.917  \\
    \bottomrule
  \end{tabular}
\end{table}

\begin{figure}[H]
  \centering
  \includegraphics[scale=0.4]{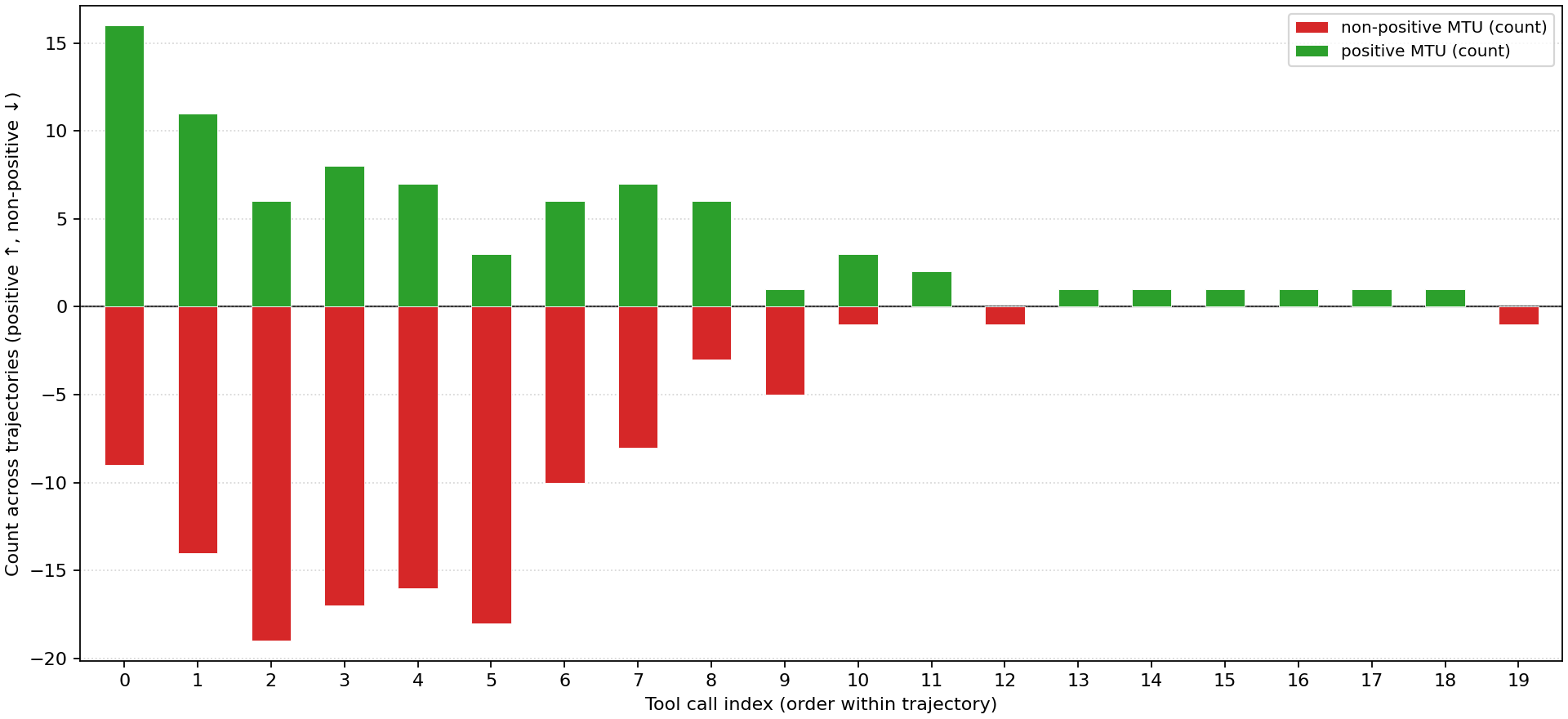}
  \caption{Marginal tool utility signs across all \texttt{default} trajectories by GPT-5.3-Codex. Different agent trajectories have different lengths, hence the reduction in total tool call count along the horizontal axis.}
  \label{fig:mtu-trend-codex}
\end{figure}

\begin{figure}[H]
  \centering
  \includegraphics[scale=0.4]{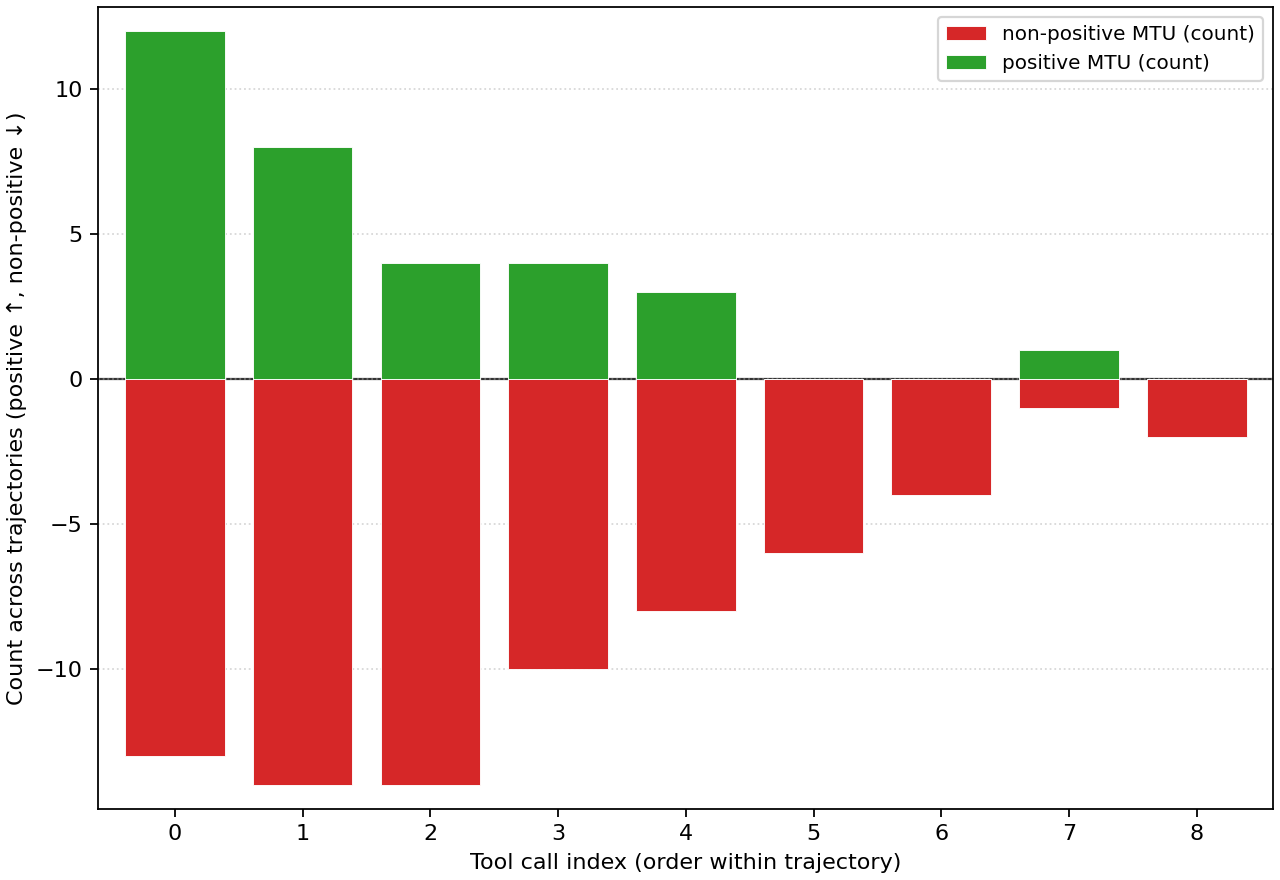}
  \caption{Marginal tool utility signs across all \texttt{default} trajectories by Gemini 3.1 Pro. Different agent trajectories have different lengths, hence the reduction in total tool call count along the horizontal axis.}
  \label{fig:mtu-trend-gemini}
\end{figure}

As seen in Figures \ref{fig:mtu-trend-codex} and \ref{fig:mtu-trend-gemini}, tool calls with $\Delta_\alpha > 0$ occur more often near the beginning of trajectories, and tool calls with $\Delta_\alpha \leq 0$ occur more often in the middle. Analyzing the agent trajectories in detail, most early calls are high-signal Grafana/Loki calls (so are most early positive marginal tool utility calls), middle calls are mostly spent querying noisy sources and performing broad conversation/specification discovery, and later calls generally surface more specific evidence to support the agent's code patch generation.

\subsection{Tool Efficiency}

Given the signs of marginal tool utility determined by the judge LLM for each tool call, we calculate tool efficiency for each trajectory in \texttt{default} and \texttt{grafana}. Then, we calculate the mean tool efficiency grouped by model and variant. Since the \texttt{no-mcp} variant has no MCP tools, tool efficiency is undefined for that particular variant. Our results are found in Tables \ref{table:teff-codex} and \ref{table:teff-gemini}.

\begin{table}[H]
  \caption{Mean tool efficiency for GPT-5.3-Codex out of 25 agent trajectories per variant. Included are the number of useful tool calls and total number of tool calls from all trajectories combined; the quotient of these two values is not mathematically equivalent to mean tool efficiency as agent trajectories vary in length.}
  \label{table:teff-codex}
  \centering
  \begin{tabular}{cccc}
    \toprule
    Variant     & Useful Call Count (All)    & Total Call Count (All) & Mean Tool Efficiency \\
    \midrule
    \texttt{default} & 82     & 204 & 0.359     \\
    \texttt{grafana} & 58     & 87 & 0.720      \\
    \bottomrule
  \end{tabular}
\end{table}

\begin{table}[H]
  \caption{Mean tool efficiency for Gemini 3.1 Pro out of 25 agent trajectories per variant. Included are the number of useful tool calls and total number of tool calls from all trajectories combined; the quotient of these two values is not mathematically equivalent to mean tool efficiency as agent trajectories vary in length.}
  \label{table:teff-gemini}
  \centering
  \begin{tabular}{cccc}
    \toprule
    Variant     & Useful Call Count (All)    & Total Call Count (All) & Mean Tool Efficiency \\
    \midrule
    \texttt{default} & 32     & 104 & 0.367     \\
    \texttt{grafana} & 23     & 43 & 0.593      \\
    \bottomrule
  \end{tabular}
\end{table}

Tool efficiency increases when Mattermost and Plane are removed. This finding is also consistent with the conclusion that these two MCP tools are unnecessarily included in the agent's tool suite and thus can be considered redundant given the nature of the task to solve.

\section{Discussion}

We identify several promising directions for future work the community can engage in that leverages marginal tool utility and tool efficiency.

\textbf{Minimizing sub-optimal middle calls.} The trend that middle calls are often sub-optimal (see Section \ref{subsection:mtu}) is a useful finding that may inspire future LLM reinforcement learning techniques, similar to those by \cite{ma2026empowering, shao2024deepseekmathpushinglimitsmathematical, ouyang2022traininglanguagemodelsfollow, uesato2022solvingmathwordproblems}, with marginal tool utility as a component of the reward function. Alternatively, marginal tool utility can be used for inference-time optimizations, like designing agents that can backtrack \citep{yang2025stepleapforwardselfbacktracking} should the trajectory thus far show some number of consecutive tool calls with non-positive marginal tool utility.

\textbf{Designing leaner tool suites.} As seen through our tool ablations of APEX-SWE Observability, tool suites can be more bloated than necessary, as has been surfaced by \cite{liu2025toolscopeenhancingllmagent} albeit from a different perspective. The tool efficiency metric can be directly used to assess the overall health of a tool suite such that it remains as lean and maintainable as possible, and the aggregate tool utility of each tool to decide whether to preserve the tool in improved versions of the harness. This enables the creation of tool suites that contain tools that serve mutually exclusive purposes. Tool descriptions and/or execution bodies \citep{lu2026statictoolstesttimetool} alone may not indicate redundancy between tools, as we have seen in Tables \ref{table:acc-codex} and \ref{table:acc-gemini}, hence marginal tool utility and tool efficiency can serve as extra sources of information to uncover such redundancies.

\textbf{More informed harness engineering.} We observe in Section \ref{section:results} that using different backbone models for agents can yield the same task accuracy with different tool efficiencies. We posit that this is primarily a product of different model training methodologies as the agent harness remains a constant variable, but an important finding to note nonetheless as this supports the view that harness engineering must be informed by each agent's particular backbone model. In other words, a one-size-fits-all approach to harness engineering may be sub-optimal, hence workflows that are designed to automate harness engineering \citep{lee2026metaharnessendtoendoptimizationmodel} can benefit from our proposed metrics.

\textbf{Self-improving agents indexing on tool efficiency.} Taking automated harness engineering one step further, the agents themselves can recursively self-improve \citep{xia2025agent0unleashingselfevolvingagents, xiong2026learningcontinuallylearnmetalearning} without relying on another agent that optimizes the harness offline \citep{cai2024largelanguagemodelstool, wolflein2025llmagentsmakingagent}. Not only can this online improvement be based on the usual metrics like task accuracy, tool efficiency and marginal tool utility can be leveraged mid-execution to provide rich reward signals to optimize context (including tool descriptions, input schemas, and output data shapes) or update model weights.

\section{Limitations} \label{section:limitations}

While confidence scores returned by an LLM as decoded tokens are not the most rigorous \citep{xiong2024llmsexpressuncertaintyempirical, yang2026calibrationlargelanguagemodels}, the judge's confidence score per classification in Tables \ref{table:mtu-codex} and \ref{table:mtu-gemini} indicate that the judge is generally more confident on non-positive classifications. We did not investigate why this is so, nor did we attempt to balance the confidence scores, as we view this work as outside the scope of this paper. Alternatively, a different LLM-as-a-Judge implementation could have been used, such as one replacing the language modeling head of the judge with a binary classification head \citep{ma2026empowering}, though this requires using an open-weight LLM.

Most notably, we only ran tool ablations for read-only tools (the three MCPs are all read-only). We did not remove tools with write privileges (e.g., apply\_patch) from the agent's tool suite as we suspect that that would have prevented the agent from completing the tasks at all. Having said that, we seek to find ways in which we can assess the marginal tool utility of write tool calls, either directly or indirectly (the latter case by investigating how read-only tool calls are affected by what was written using the write tool calls), in future experiments.

\section{Conclusion}

We introduce tool efficiency and marginal tool utility, both being new quantitative metrics that are relevant in determining the usefulness of tools and thus in constructing the leanest possible tool suite for an LLM agent given a particular task. Our findings suggest that intuitive judgments regarding a tool's usefulness as well as empirical tool ablation results using accuracy as a proxy in discerning usefulness/necessity of different tools agree with conclusions drawn from analyzing marginal tool utility and tool efficiency values.

All in all, we look forward to what our new definitions enable in the development of LLMs and LLM agents. We hope that the literature matures not only in terms of new methods, architectures, and workflows, but also in new evaluation dimensions against which we can measure the holistic quality of these new artifacts. As well-defined quantitative metrics are more tractable to optimize for, we believe that evaluation/metrics research will accelerate progress in LLM research and engineering further, providing the necessary direction and focus guiding iteration cycles of novel methods.

\textbf{Societal Impacts.} As large language models and agents built on them become more capable, many have raised concerns regarding our over-reliance, even dependence, on them. While this work does not introduce a new artifact directly relevant to such concerns, this work does enable new artifacts to be more capable than their existing versions today. We want to acknowledge that, and it is our hope and belief that agents and humans are able to work in tandem, each tackling a dimension of work the other is less-suited for. The onus is on us, those at the frontier of the development of this technology, to understand and thus educate the public on what those dimensions are.

\begin{ack}
All funding for this work comes from Foam.
\end{ack}

\bibliography{references}



\appendix

\section{Prompt for LLM-As-A-Judge for Marginal Tool Utility Sign} \label{app:judge-mtu}

$\newline$

\begin{tcolorbox}[
  enhanced,
  breakable,
  colback=white,
  colframe=gray!70!black,
  boxrule=1.2pt,
  arc=4mm,
  left=8mm,
  right=8mm,
  top=8mm,
  bottom=6mm,
  before skip=6pt,
  after skip=10pt,
  overlay unbroken and first={
    \node[
      fill=gray!15,
      draw=gray!70!black,
      line width=1.2pt,
      rounded corners=2mm,
      inner xsep=18mm,
      inner ysep=2mm
    ] at ([yshift=0mm]frame.north) {System Prompt};
  }
]

You are an expert evaluator of agent trajectories completing a SWE bugfix observability task.\\

Your task is to determine whether a tool call increases the likelihood the bugfix task is solved correctly. This is measured by a metric called marginal tool utility (MTU).\\

Concretely, you are tasked to determine the sign of marginal tool utility (MTU) for one tool call.\\

Definition:\\
MTU(i) = p(task solved correctly | tool calls 0..i) - p(task solved correctly | tool calls 0..i-1)\\

Equivalently:\\
MTU(i) is positive if the likelihood of the bugfix task being solved correctly strictly increases after tool call i is added into the trajectory. MTU(i) is non-positive otherwise.\\

You are given:
\begin{itemize}
    \item BEFORE context: state of trajectory before tool call i executes.
    \item AFTER context: state of trajectory after tool call i completes.
    \item The specific tool call and tool result.\\
\end{itemize}

Output strict JSON only:\\
\{\\
    "label": "positive" | "non\_positive",\\
    "confidence": float between 0 and 1,\\
    "rationale": "brief explanation"\\
\}\\

Use "positive" only when the call clearly increases solve likelihood.\\
Use "non\_positive" when likelihood is unchanged or reduced.

\end{tcolorbox}

$\newline$

\begin{tcolorbox}[
  enhanced,
  breakable,
  colback=white,
  colframe=gray!70!black,
  boxrule=1.2pt,
  arc=4mm,
  left=8mm,
  right=8mm,
  top=8mm,
  bottom=6mm,
  before skip=6pt,
  after skip=10pt,
  overlay unbroken and first={
    \node[
      fill=gray!15,
      draw=gray!70!black,
      line width=1.2pt,
      rounded corners=2mm,
      inner xsep=18mm,
      inner ysep=2mm
    ] at ([yshift=0mm]frame.north) {User Prompt};
  }
]

Decide MTU sign for this tool call.\\

Target tool call:
\begin{itemize}
    \item tool\_call\_id: \{TOOL\_CALL\_ID\}
    \item tool\_name: \{TOOL\_NAME\}
    \item arguments: \{ARGUMENTS\_TRUNCATED\}
    \item tool\_result\_excerpt: \{TOOL\_RESULT\}\\
\end{itemize}

BEFORE context (through tool calls 0..i-1 completion):\\
=== BEFORE START ===\\
\{BEFORE\_CONTEXT\}\\
=== BEFORE END ===\\

AFTER context (through tool calls 0..i completion):\\
=== AFTER START ===\\
\{AFTER\_CONTEXT\}\\
=== AFTER END ===\\

Return strict JSON only.

\end{tcolorbox}

\section{APEX-SWE Observability Per-Task Correctness Breakdown} \label{app:acc-breakdown}

\begin{figure}[H]
  \centering
  \includegraphics[scale=0.135]{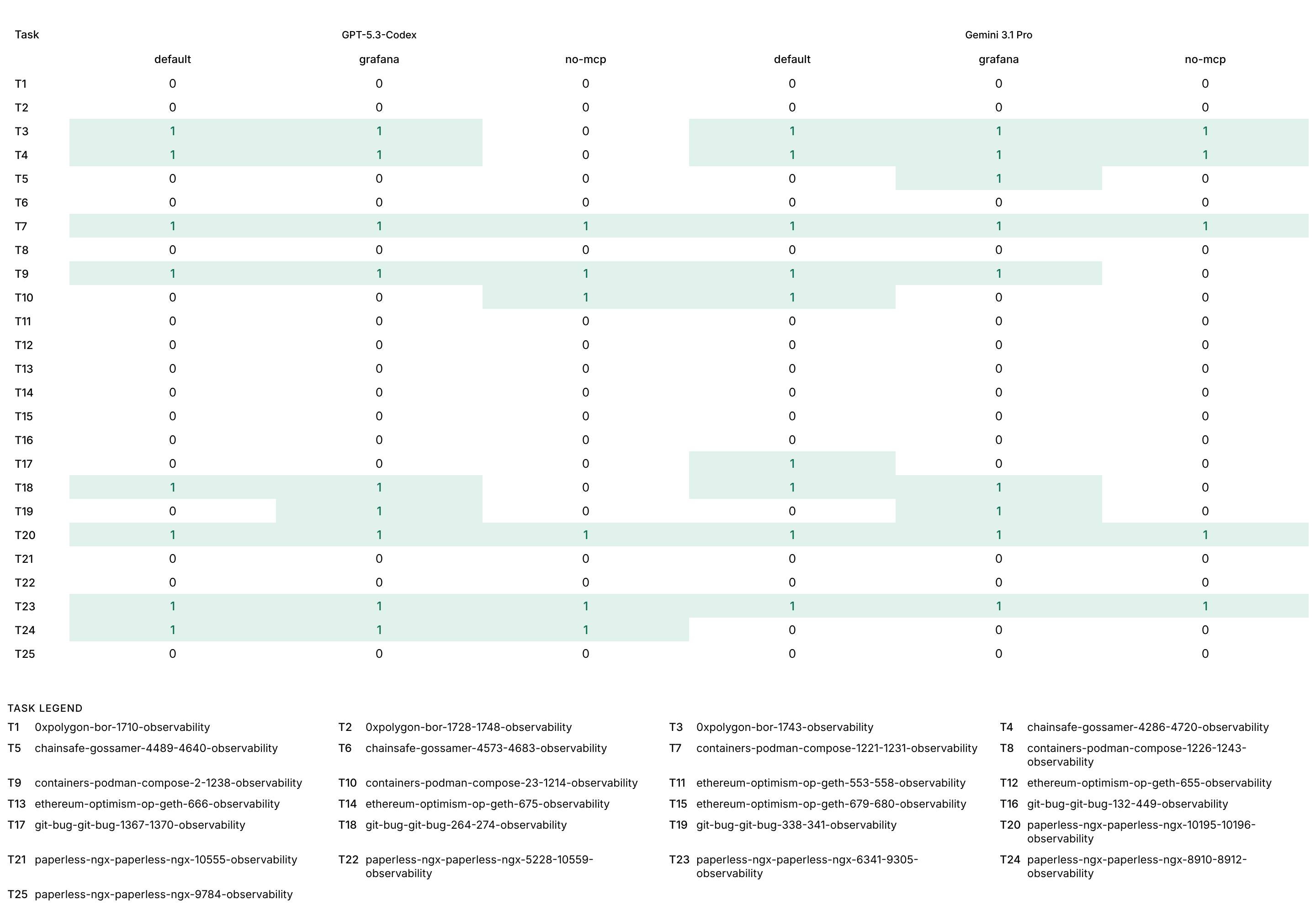}
  \caption{Full score breakdown per task. A score of $1$ indicates a pass, while a score of $0$ indicates a fail.}
  \label{fig:scores}
\end{figure}





\end{document}